# Deep Learning Based Regression and Multi-class Models for Acute Oral Toxicity Prediction with Automatic Chemical Feature Extraction

Youjun Xu,[†] Jianfeng Pei,[*,†] and Luhua Lai[*,†,‡,¶]

†Center for Quantitative Biology, Academy for Advanced Interdisciplinary Studies, Peking University, Beijing 100871, China

‡Beijing National Laboratory for Molecular Sciences, State Key Laboratory for Structural Chemistry of Unstable and Stable Species, College of Chemistry and Molecular Engineering, Peking University, Beijing 100871, China

¶Peking-Tsinghua Center for Life Sciences, Peking University, Beijing 100871, China

E-mail: jfpei@pku.edu.cn; lhlai@pku.edu.cn

Fax: (+86)10-62759595; (+86)10-62751725

## Abstract

Median lethal death, $LD_{50}$, is a general indicator of compound acute oral toxicity (AOT). Various *in silico* methods were developed for AOT prediction to reduce costs and time. In this study, we developed an improved molecular graph encoding convolutional neural networks (MGE-CNN) framework to construct three types of high-quality AOT models: regression model (deepAOT-R), multi-classification model (deepAOT-C) and multi-task (deepAOT-CR). These predictive models highly outperformed previously reported models. For the two external datasets containing 1673 (test set I) and 375 (test set II) compounds, the $R^2$ and mean absolute error (MAE) of deepAOT-R on the test set I were 0.864 and 0.195, and the prediction accuracy of deepAOT-C was 95.5% and 96.3% on the test set I and II, respectively. The two external prediction accuracy of deepAOT-CR is 95.0% and 94.1%, while the $R^2$ and MAE are 0.861 and 0.204 for test set I, respectively. We then performed forward and backward exploration of deepAOT models for deep fingerprints, which could support shallow machine learning methods more efficiently than traditional fingerprints or descriptors. We further performed automatic feature learning, a key essence of deep learning, to map the corresponding activation values into fragment space and derive AOT-related chemical substructures by reverse mining of the features. Our deep learning framework for AOT is generally applicable in predicting and exploring other toxicity or property endpoints of chemical compounds. The two deepAOT models are freely available at `http://www.pkumdl.cn/DLAOT/DLAOThome.php`.

## Introduction

Evaluating chemical acute toxicity is important in avoiding potential harmful effects of compounds on human health. $LD_{50}$, the dose of a chemical that causes a 50% death rate in test animals after administration of a single dose,[1] is a general indicator used to measure the acute toxicity of a compound. *In vivo* experiments of animal tests are required to accurately determine acute chemical toxicity, although these procedures are complicated, costly, and time-consuming. In addition, due to animal rights, $LD_{50}$ testing of animals is highly controversial.[2] Therefore, new reliable *in silico* methods need to be developed in comparison to standard in vivo experiments in predicting chemical acute toxicity.

Currently, many quantitative structure-property relationship (QSPR) models have been developed to predict acute rodent toxicity of organic chemicals. In these studies, there are various mathematical methods applied to construct regression models (RMs) and classification models (CMs), such as multiple linear regression (MLR),[3–6] linear regression,[7,8] neural network (NN),[9–12] $k$ nearest neighbors,[13,14] random forest (RF),[13,14] hierarchical clustering,[13] support vector machine (SVM),[14,15] relevance vector machine (RVM),[14] and local lazy learning (LLL).[16] In terms of RMs, Lu *et al.*[16] constructed prediction models using the LLL method, which yielded a maximized linear correlation coefficient ($R^2$) for large test sets. The $R^2$ of consensus RM based on LLL was 0.608 for "Set_3874". Lei *et al.*[14] argued that this method relies on prior knowledge of the query neighbour experimental data, such that the actual prediction capability was associated with the chemical diversity and structural coverage of the training set. However, machine learning methods have demonstrated potential in establishing complex QSPRs for data sets that contain diverse ranges of molecular structures and mechanisms. Thus, Lei *et al.* employed RVM combined with other methods ($k$ nearest neighbor, RF, SVM *etc.*) to construct a consensus RM for predicting AOT in rat. The predictive $R^2$ for the external test set (of 2736 compounds) was 0.69. Li *et al.*[15] suggested that a multi-classification model (MCM) might be more intuitive in toxicity estimation than a regression model (RM), as a toxic classification is easier to interpret. According to the classification criterion of the U.S. Environmental Protection Agency (EPA) (category I: (0, 50]; category II: (50, 500]; category III: (500, 5000]; category IV: (5000, $+\infty$); mg/kg), MCM with one-vs-one ($SVM_{OAO}$) and binary tree SVM methods were developed based on different molecular fingerprints or descriptors, yielding an accuracy of 83.2% for validation set (2049 compounds), 83.0% for test set I (1678 compounds), and 89.9% for test set II (375 compounds). In chemoinformatics research, high-quality QSPR models with interpretable relationship between chemical properties and chemical features are especially welcome. However, predictive power and interpretability of QSPR models are two different objectives that are difficult to achieve simultaneously.[17] We can identify some important features from weights within linear-based models (MLR, linear-SVM *etc.*) with low predictive power. These features may be mapped into the corresponding fragments in chemical structural space. With statistics and sensitivity analysis[18] of input features (rather than intuitive analysis of the constructed models), "black box" models (NN, kernal-SVM, RF) with high predictive power can extract human understandable knowledge. These above methods depend on complicated molecular representation (MR) using chemical knowledge and intuition.

Appropriate MRs that are related to biological activities or other experimental endpoints[19,20] are crucial in developing accurate prediction models. Automatic representation would greatly simplify and accelerate the development of QSPR models. The emergence of deep learning techniques[21–23] may provide possible solutions to this problem. Instead of using application-specific molecular descriptors or fingerprints (*e.g.* ECFP,[24] MACCS,[25] *etc.*), the AOT issue can be resolved using raw and pertinent features without manual intervention or selection. The two-dimensional (2D) structure of a small molecule is equivalent to an undirected graph, with atoms as nodes and bonds as edges. Encoding an undirected graph can be converted into a problem of fitting a graph into a fix-sized vector. Currently, two types of methods, sink-based and source-based, have been used for encoding undirected graphs with NNs. In the sink-based method, by defining a root node, all the other nodes in the graph proceed towards the root. The internal process is embedded with multiple NNs in representing the information transmission between nodes, after which the final information is extracted from the root node. The sink-based method was demonstrated to be feasible and practical.[26,27] However, there are no reasonable explanation for hidden-layer features in deep learning model such that the model seems "black". In the source-based method, similar to the Morgan algorithm[28] and extended-connectivity fingerprints (ECFP),[24] when starting from an initial node and diffusing outward layer-by-layer with multiple NNs, the information can be extracted step-wise from each layer. Recently, Duvenaud *et al.*[29] and Kearnes *et al.*[30] first used CNNs to successfully implement similar source-based methods. The state-of-the-art performance on some public datasets[31–36] suggests that molecular graph encoding (MGE) methods based on multiple NNs have potential in the field of chemoinformatics. In principal, MGE is an ideal representation of chemical structures without information loss.

Actually, intermediate features within deep learning models are far from random, uninterpretable patterns. By visualizing the activity of hidden layers based on well-performed models from ImageNet 2012, Zeiler *et al.* presented a nested hierarchy of concepts, with each concept defined in relation to simpler concepts (pixels → edges → corners and contours → object parts → object identity),[37] which is an efficient illustration of a deep learning-based CNN model. Different compounds may play different functions in the living organisms. Simple concepts of atoms and bonds are combined into more complex concepts of structural fragments, then integrated into high concepts of different functions (atoms and bonds→ fragments → functions). By designing ECFP-based CNN architecture, the internal features were visualized by Duvenaud *et al.* as the corresponding fragments,[29] providing a better understanding of a deep learning-based QSPR model. Despite of a number of successful application examples in chemoinformatics studies using MGE, the following points need to be improved for better prediction and easy interpretation: 1) hyperparameters, 2) training and prediction strategy, 3) multi-output problem, 4) model interpretation. The approach based on CNN with these above improvements was referenced hereafter as "MGE-CNN".

Here we used MGE-CNN framework (shown in Figure 1A) to construct AOT models. In order to develop high-quality deep learning models, namely deepAOT, RMs were constructed using the reported largest AOT dataset from Li *et al.*,[15] including experimental oral $LD_{50}$ values for chemicals in rat. Based on the U.S. EPA criterion for the AOT category, MCMs were also developed to predict chemical toxicity categories. Two external test datasets were used to estimate the predictive power of RMs and MCMs. The consensus RM and the best MCM were called "deepAOT-R" and "deepAOT-C", respectively. We demonstrated that the deepAOT-R and deepAOT-C models outperformed the previous reported models whether it was a regression or classification problem. Given the relevance of both tasks, multi-task deepAOT-CR model was developed for improving the consistency of regression and classification models. Further analysis was performed by forward and backward exploration (Figure 1A) of internal features (referred to as deep fingerprints) directly extracted from our models to interpret the RMs and MCMs. The forward exploration was used to determine the predictability of fingerprints, while the backward exploration was used to understand and explore structural alerts concerning AOT. In view of end-to-end learning, the MGE-CNN framework in this study can also be applied to predicting and exploring other toxicity endpoints induced by small molecules in complex systems.

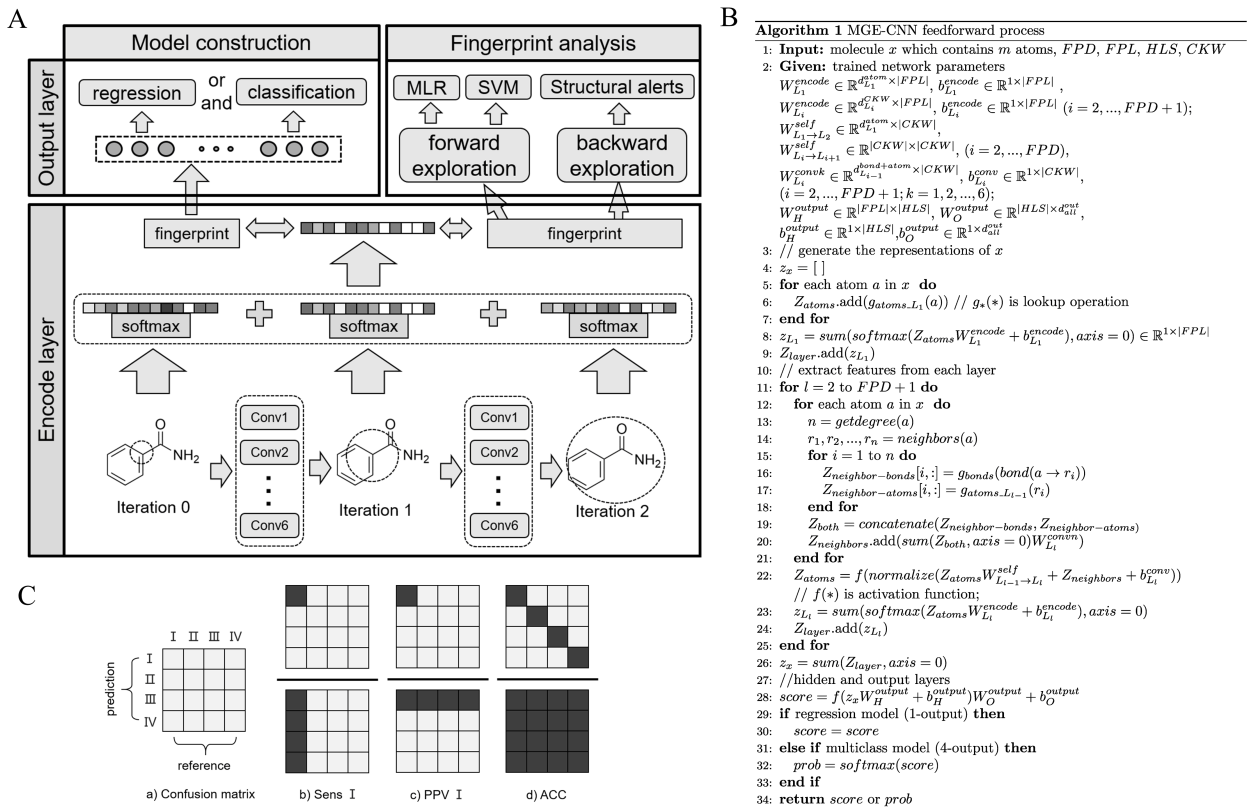

Figure 1: (A) Schematic diagram of MGE-CNN architecture. "Conv" represents the convolution kernel and the 6 kernels rely on the degree of each atom. (B) Overview of pseudocode in Algorithm 1. (C) The assessment method of Sens and PPV for each of classes and ACC of all the classes. Sens I is equal to the number of the higher black region divided by the sum of the bottom black region, which was identical with PPV I. The roman letters "I, II, III, IV" represent toxicity categories.

## Materials and Methods

### MGE-CNN

The MGE-CNN architecture takes the canonical SMILES string of a small molecule as input, and produces a score capable of describing a value or label about toxicity. Figure 1A and 1B show this architecture and its high-level pseudocode with the steps of MGE-CNN feedforward process. Firstly, given an input SMILES string ($x$), a molecular structural graph is converted by the RDKit toolbox.[38] The sub-graph from each layer (or iteration) is encoded into a fixed-sized vector $z_{L_l} \in \mathbb{R}^{|FPL|}$  $l \in \{1, 2, ..., |FPD|\}$, then these vectors are summed as $z_x \in \mathbb{R}^{|FPL|}$ representing this molecule. Then $z_x$ is used as input of the subsequent neural network in the output layer for executing the following operation:

$$score = f(z_x W_H^{output} + b_H^{output}) W_O^{output} + b_O^{output} \qquad (1)$$

where $W_H^{output} \in \mathbb{R}^{|FPL| \times |HLS|}$ is the weight matrix of hidden layer in the output layer, $W_O^{output} \in \mathbb{R}^{|HLS| \times d_{all}^{out}}$ is the weight matrix of output layer in the output layer, and $b_H^{output} \in \mathbb{R}^{1 \times |HLS|}$ and $b_O^{output} \in \mathbb{R}^{1 \times d_{all}^{out}}$ are bias terms. $d_{all}^{out} = 1$ for RMs, $d_{all}^{out} = 4$ for MCMs. The 4-dimensional vector is transformed with $softmax$ function representing the probability of four classes. $p(i|x) = \frac{e^{score(x)_i}}{\sum_{j=1}^{4} e^{score(x)_j}}$ is the probability of category $i$, where $score(x)_i$ is the score for category $i$.

The MGE-CNN has three main advantages: 1) The input information of initial atoms and bonds is very similar to that of ECFP. The atom information contains atomic type, its degree, its implicit valence, the number of attached H atoms and aromatic atoms. The bond information is relied on bond type (single, double, triple, aromatic, conjugated or in-a-ring). These atom and bond-level information is used to characterize the surrounding chemical environment of each atom as completely as possible. All of these information can be calculated using RDKit. 2) Molecular graphs are encoded with CNN, which makes information transmission become continuous and constructs an end-to-end differential system. In such case, we can perform gradient descent with a large number of labelled data to optimize this system. During the training process, automatic feature learning is implemented, avoiding manual feature selection. 3) The feature learning and model construction processes are integrated together. Once the model is well-trained with supervised learning, these fingerprints are also learned.

The following improvements for better prediction and easy interpretation in our system were adopted: 1) For hyperparameter optimization in the AOT system, we empirically found that the default settings ($\beta_1 = 0.9, \beta_2 = 0.999$) for adaptive moment estimation (Adam) would be more helpful than those provided by the Duvenaud *et al.* 2) To avoid providing the training examples in a meaningful order (which may bias the optimization algorithm and lead to over-fitting), the trick of "shuffling"[39] was added into the whole training process. 3) The popular methods of *softmax* function and *cross-entropy* loss function were introduced to meet the requirements of multi-classification task. 4) Regression and classification tasks were taken into consideration simultaneously for developing the multi-task model. 5) To further explain the rationality of our models, deep fingerprints directly extracted from well-built models were used to construct shallow machine learning models. The structural fragments with the largest contribution ($\arg\min$ (linear regression coefficient × activation values) ) to chemical toxicity were drawn out for comparison with the reported toxicity alerts, while the original MGE only considered those coefficients. 6) The mean and standard deviation of the training set for each layer are calculated for normalizing validation or external test set, reducing the bias caused by different distributions. Based on these, the MGE-CNN was employed to construct RMs and CMs for estimating AOT in rat, as shown in Figure 1A. During "Model construction", these models were trained, validated and externally challenged. During "Fingerprint analysis", the well-trained deep fingerprints of small molecules were used to develop shallow models, MLR and SVM, to predict AOT values or labels. Simultaneously, the most relevant feature among deep fingerprint for each compound was calculated based on linear regression with least squares fitting, then traced back to the atomic level, and mapped onto AOT activation fragments. These activated fragments were then used to compare with reported toxicity alerts (TAs) to validate the inference capability for TAs.

## Training deepAOT models

The approach for training deepAOT models includes hyperparameter optimization methods and gradient descent optimization algorithms.

**Hyperparameter optimization**

Deep learning is a dramatic improvement in many fields,[21] in particular for CNNs,[40–42] which are often able to automatically learn useful features with little manual intervention of data through multiple layers of abstraction. However, these successes do not detract from the advantages of hyperparameter optimization. An appropriate set of hyperparameters must be selected before applying deep learning framework for a new data set, which is a time-consuming and tedious task.[43] The hyperparameters of MGE-CNN include the length of fingerprint (FPL), the depth of fingerprint (FPD), the width of convolution kernel (CKW), the size of hidden units in the output layer (HLS), the $L2$ penalty of cost function (L2P), the scale of initial weights (IWS) and the step size of learning rate (LRS). The ranges of these parameters are shown in Table S1, as recommended by Duvenaud *et al.*(`github.com/HIPS/neural-fingerprint/issues/2`) In order to reduce computational costs, a simplified parameter range was used as follows: FPL $\in \{16, 32, 48, 64, 80, 96, 112, 128\}$; FPD $\in \{1, 2, 3, 4\}$; CKW $\in \{5, 10, 15, 20\}$; HLS $\in \{50, 60, 70, 80, 90, 100\}$; $log$(L2P) $\in \{-6, -5, -4, -3, -2, -1, 0, 1, 2\}$; $log$(IWS) $\in \{-6, -5, -4, -3, -2\}$; $log$(LRS) $\in \{-8, -7, -6, -5, -4\}$.

Usually, the three most popularly used methods for hyperparameter optimization are manual search, grid search, and random search. Of these methods, random search was demonstrated to outperform a combination of manual and grid search when applied to a set of problems.[44] Therefore, random search was used to generate 500 sets of hyper-parameters for RMs and CMs and all hyperparameter sets were evaluated with the validation set (2045 compounds). The top 10 models were then applied to the next step in selecting the model with lowest root mean square error (RMSE) for RMs, eventually selecting models with the highest accuracy (ACC) for MCMs.

**Gradient descent optimization**

Gradient descent is one of the most popular algorithms to optimize deep learning-based networks. Every state-of-the-art deep learning library contains implementations of various algorithms to optimize gradient descent.[45] Adaptive Moment Estimation (Adam)[46] is a popular method that computes adaptive learning rates for each weight. It takes an exponentially decaying average of past gradients and past squared gradients into consideration and demonstrates empirically that Adam works well for adaptive learning-method algorithms. The shuffle of training set after each epoch was also applied to the training process for avoiding bias of the optimization algorithm. Therefore, the training strategy was implemented by a pseudocode of Algorithm 2 in Supporting Information.

$$J(\theta) = \frac{1}{n} \sum_{i=1}^{n} (\hat{y}_i - y_i)^2 + \alpha \|\theta\|_2 \tag{2}$$

$$J(\theta) = -\frac{1}{n} \left[ \sum_{i=1}^{n} \sum_{j=1}^{k} 1 \left\{ y^{(i)} = j \right\} \log \frac{e^{\theta_j^T x^{(i)}}}{\sum_{l=1}^{k} e^{\theta_l^T x^{(i)}}} \right] + \alpha \|\theta\|_2 \tag{3}$$

where $J(\theta)$ is the loss function added $L2$ penalty described in Equations 2 & 3, which were used to evaluate RMs and MCMs, respectively. A flexible automatic differentiation package called Autograd (`https://github.com/HIPS/autograd`) was easily adopted for computing gradients of weights.

## Experimental Setup

### Data Collection and Preparation

The AOT database provided by Li *et al.*,[15] the largest data set for oral $LD_{50}$ in rat, was used in this study. All data was from three sources: 1) the admetSAR database;[47] 2) the MDL Toxicity Database (version 2004.1),[48] and 3) the Toxicity Estimation Software Tool (TEST version 4.1)[49] program from the U.S. EPA. The preparation of the data set had been executed by Li *et al.*[15] The "Structure Checker" and "Standardizer" modules from the ChemAxon Inc. (evaluation version)[50] were used to fix some error valence and standardize all the SMILEs strings in the dataset. The workflow is shown in Figure S1. Finally, the training and validation sets included 8080 and 2045 compounds, respectively, with measured

LD$_{50}$ values adopted from the admetSAR database. Two external data sets contained 1673 (from MDL Toxicity Database) and 375 (from TEST) compounds. Based on the U.S. EPA definition of toxicity,[51] all compounds were divided into four categories based on their levels of toxicity. The statistical description of the entire data set is shown in Table 1. The entire data set was consistent with observations made by Li *et al.*'s (training set: 8102; validation set: 2049; test set I: 1678; test set II: 375). Test set II only had category labels without exact experimental values of acute oral LD$_{50}$.

Table 1: Statistical description of the training, validation, and external test sets.

| Category | I | II | III | IV | Total |
|---|---|---|---|---|---|
| Training set | 794 | 1933 | 4303 | 1050 | 8080 |
| Validation set | 224 | 463 | 1155 | 203 | 2045 |
| Test set I | 92 | 341 | 1099 | 141 | 1673 |
| Test set II | 57 | 93 | 183 | 42 | 375 |
| Total | 1167 | 2830 | 6740 | 1436 | 12173 |

**Construction Strategy of RMs and MCMs**

RMs and MCMs were constructed by MGE-CNN. For RMs, the training target was a $log$(LD$_{50}$) (unit: $log$(mg/kg)) value for each compound. The loss function of Equation 2 was adopted in the MGE-CNN. In order to select appropriate sets of hyperparameters, each set of 500 random combinations was run for 750 epochs with a mini-batch gradient descent and Adam optimization algorithm. We selected the top 10 sets of hyperparameters with lowest RMSE values of the validation set. Generally, the purpose of 10 well-trained models is to quantitatively predict $log$(LD$_{50}$) of unknown compounds. Therefore, the 10 models needed to be challenged by an external data set (Test set I) (note: test set II lacks the LD$_{50}$ values). The consensus RM (deepAOT-R) was constructed with averaging the previous 10 models and the classification capacity of the deepAOT-R model was estimated and analyzed.

For MCMs, the training target was a defined label of compound toxicity. According to the category criterion, four categories also meant four outputs in the MGE-CNN architecture. The $softmax$ loss function (Equation 3) was used as the object function for MCMs. Initially, each of the 500 random sets of hyperparameters was run for 1000 epochs to select the top 10 sets with highest ACC of the validation set. Next, the top 10 models were run for an additional 1000 epochs. Finally, the best-trained weights were selected out with the highest ACC of the validation set. Consequently, the best 10 MCMs were challenged by the two external test sets. Meanwhile, the consistency between MCMs and RMs was analyzed according to their prediction outcomes.

**Forward and backward exploration of Fingerprints**

In order to determine what these models actually predict, the forward and backward exploration approach was applied for "Fingerprint" layer. The forward exploration was implemented by extracting the values of "Fingerprint" layer (deep fingerprints) to construct MLR and SVM models. This could demonstrate the support degree that these features provided in the shallow machine learning decision-making system. While assessing the performance of shallow models with deep fingerprints, increased performance would suggest optimized predictive features from this MGE-CNN architecture.

The backward exploration is that after linear regression, the most linear-negative-correlation feature was selected from the $|FPL|$-dimensional "Fingerprint" layer. Further analysis examined that related atoms and their neighboring atoms, with the most prominent contribution to this feature were reversely calculated out, which was called activation fragment. The activation fragment is highlighted in a drawing of each compound presented in category I. These highlighted fragments were considered by prediction models to be substructures most related to AOT, which an inference to toxicity fragments. Meanwhile, these fragments were used to make comparisons with the reported structural features from the Online Chemical Database (ToxAlerts)[52] for validating the inference capability of MGE-CNN-based deepAOT models.

## Evaluation Metrics

All of the models were evaluated using the validation set, then challenged by two external test sets. The three indexes of RMSE (Equation 4), mean absolute error (MAE, Equation 5) and square of Pearson correlation coefficient (PCC$^2$, Equation 6) were used as evaluation indexes for the RMs. The MCMs were assessed in accordance with the multi-class confusion matrix, in which the sensitivity (Sens), positive predictive value (PPV), and ACC were calculated as shown in Figure 1C. In addition, the consensus deepAOT-R model was used to assess classification performance. The PCC is a description of linear correlation and a regression line estimates the average value of target $y$ for each value of input $X$, but actual $y$ values differ from the predicted values $\hat{y}$ unless the correlation is perfect. These differences are called prediction errors or residuals, which means that it is reasonable and valuable for a predicted value accomplished by a wiggle room to judge this prediction. Thus, 1-fold RMSE for the validation set was added into the outcomes of RMs. For the two external test sets, deepAOT-R predicted the output values, which were then mapped into the category space and transformed into the output labels. The ranges of output labels were calculated with the output values within 1 RMSE. Assuming that the range of a predicted label contains the actual target label, this prediction was considered to be correct.

$$RMSE = \sqrt{\frac{\sum_{i=1}^{n}(\hat{y}_i - y_i)^2}{n}} \tag{4}$$

$$MAE = \frac{1}{n}\sum_{i=1}^{n}|\hat{y}_i - y_i| \tag{5}$$

$$PCC^2 = \left[\frac{\sum_{i=1}^{n}(x_i - \bar{x})(y_i - \bar{y})}{\sqrt{\sum_{i=1}^{n}(x_i - \bar{x})^2}\sqrt{\sum_{i=1}^{n}(y_i - \bar{y})^2}}\right]^2 \tag{6}$$

# Results and Discussion

## Performance Evaluation of RMs

The RMs help to quantitatively predict the $log(\text{LD}_{50})$ values in rat for compounds, reflecting their toxicity: the smaller the value, the more toxic the compound. The 500 random sets of hyperparameters were fed into the MGE-CNN architecutre and those 500 models were trained with different hyperparameters for 750 iterations to construct the RMs.

The RMSE and PCC indexes of the training and validation sets from 500 models after gradient-based optimization training were shown in Figure S2. For the training and validation sets, decreased RMSE was accompanied by a progressive increase of PCC, which completely conformed to the logical law of gradient descent. The three indexes of RMSE, MAE and $\text{PCC}^2$ over 500 models with different hyperparameters had a wide range of changes and the whole performance of the top 10 RMs is shown in Table S2 and Figure 2A, in which MAE, RMSE and $\text{PCC}^2$ on the three sets are described. Among the 10 RMs, RM4 had the best MAE (0.287), RMSE (0.382), $\text{PCC}^2$ (0.804) for the training set, but a sub-optimal performance for the validation set (MAE of 0.258, RMSE of 0.337, $\text{PCC}^2$ of 0.867). For test set I, RM4 also has the optimal performance of 0.245 for MAE, 0.319 for RMSE, 0.804 for $\text{PCC}^2$. The consensus outcomes display a further improvement of the three indexes for the three data sets. For example, $\text{PCC}^2$ was 0.853 for the training set (with a 0.049 increase), 0.917 for the validation set (with a 0.037 increase) and 0.864 for test set I (with a 0.060 increase). These deepAOT-R outcomes outperformed the consensus model from Lei *et al.*[14] (0.487 for MAE, 0.646 for RMSE, 0.690 for $\text{PCC}^2$). The distribution of prediction errors (predictions - targets) for the three sets is shown in Figure S3, which was a reasonable distribution for training and prediction results. Therefore, it was necessary for the MGE-CNN architecture to optimize hyperparameters, which would help to boost the performance. Moreover, the ensemble strategy demonstrated that the deepAOT-R had the optimal performance.

In order to investigate classification abilities of the RMs, the consensus model, deepAOT-

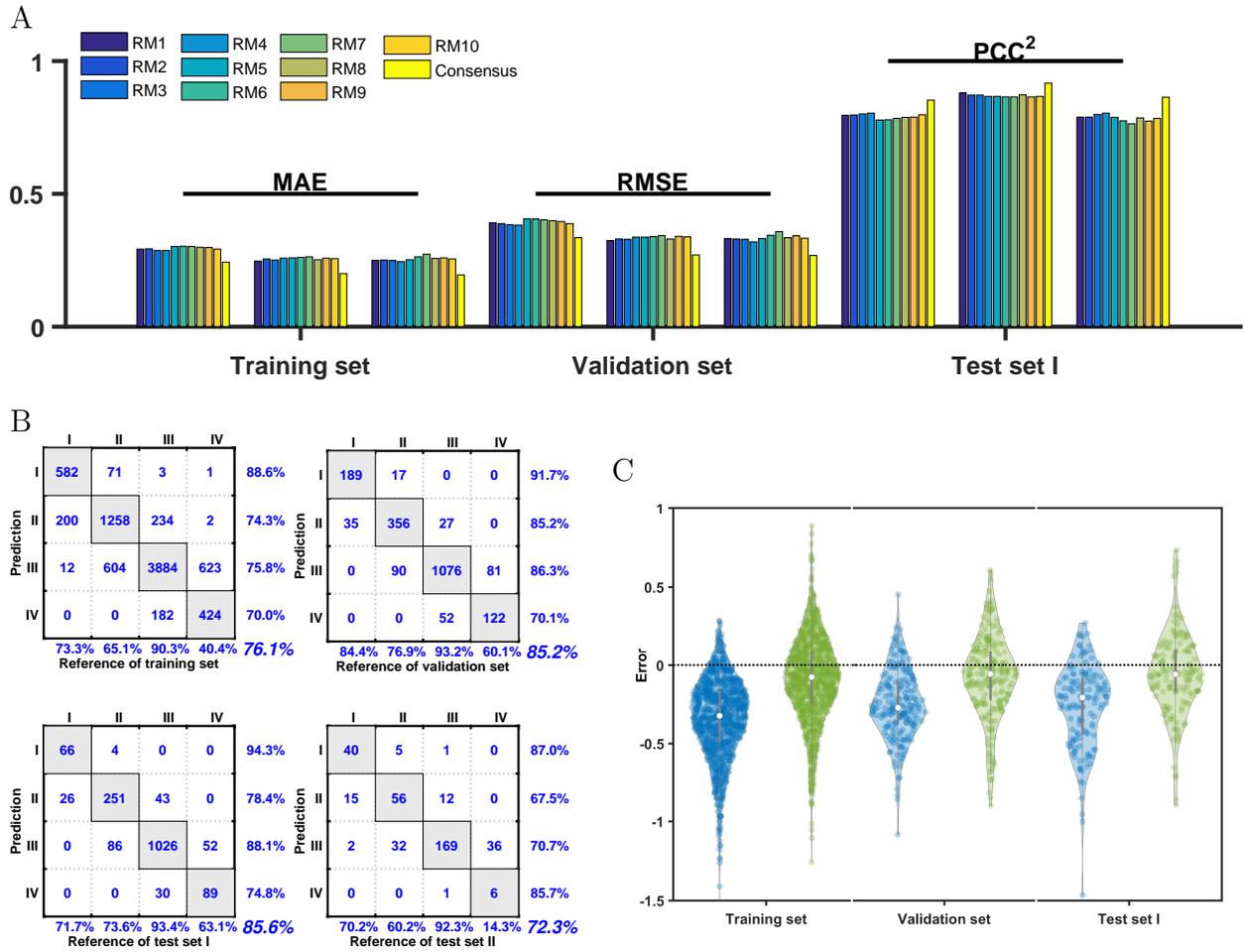

Figure 2: Performance overview of the top 10 RMs and the consensus deepAOT-R model. (A) The overview of MAE, RMSE and PCC$^2$ index for all the RMs. (B) The confusion matrix for assessing deepAOT-R's classification capacity. (C) The distribution comparison of regression prediction errors from category IV. Blue color: deepAOT-R; Green color: deepAOT-CR.

R, was used to predict the toxicity labels for all of the data sets (test set II had toxicity labels, but lacked $LD_{50}$ values). The predicted values $log(LD_{50})$ were transformed into $LD_{50}$ values and mapped into category space, then the multiclass confusion matrix is summarized in Figure 2B, where the Sens, PPV, and ACC index for each class are shown at the bottom of the box, the right of the box, and as a number in the bottom right corner, respectively. The overall performance was at an acceptable level, although there were poor levels among the four sets when examining the Sens IV index, dividing the compounds with category IV into category III, which suggested that deepAOT-R could not distinguish well between category III and IV. The prediction error distribution of category IV is presented in Figure 2C (in blue), suggesting that most prediction errors of category IV were lower than zero and might cause such phenomena. However, when the 1-fold ValRMSE (0.270) wiggle room was taken into consideration, the classification performance significantly improved (Figure S4), which revealed that the deepAOT-R outcomes were still relatively close to the actual target values. Hence, deepAOT-R had a certain distinguishing power of classification, indicating that a wiggle room of 1-fold ValRMSE could be useful for prediction results.

Table 2: Hyper-parameters and performance of the top 10 multi-classification models.

| Model | | CM1 | CM2 | CM3 | CM4 | CM5 | CM6 | CM7 | CM8 | CM9 | CM10 |
|---|---|---|---|---|---|---|---|---|---|---|---|
| Hyper-param | FPL | 80 | 128 | 128 | 48 | 112 | 48 | 16 | 128 | 16 | 112 |
| | FPD | 4 | 2 | 3 | 2 | 3 | 3 | 3 | 4 | 3 | 2 |
| | CKW | 20 | 20 | 20 | 20 | 20 | 20 | 15 | 15 | 20 | 15 |
| | HLS | 90 | 60 | 90 | 70 | 50 | 60 | 50 | 50 | 70 | 90 |
| | $log$(L2P) | -4 | -5 | -3 | -3 | -4 | -2 | -4 | -6 | -3 | -5 |
| | $log$(IWS) | -1 | -1 | -1 | -1 | -1 | -1 | -1 | -1 | -1 | -1 |
| | $log$(LRS) | -4 | -4 | -4 | -4 | -4 | -5 | -4 | -5 | -5 | -4 |
| Evaluation index* | preTrainACC | 0.902 | 0.866 | 0.902 | 0.802 | 0.891 | 0.788 | 0.764 | 0.810 | 0.768 | 0.810 |
| | preValACC | 0.940 | 0.914 | 0.942 | 0.869 | 0.941 | 0.881 | 0.845 | 0.883 | 0.839 | 0.880 |
| | TrainACC | 0.921 | 0.920 | 0.908 | 0.841 | 0.934 | 0.802 | 0.790 | 0.855 | 0.812 | 0.881 |
| | ValACC | 0.958 | 0.959 | 0.942 | 0.905 | 0.963 | 0.891 | 0.866 | 0.922 | 0.887 | 0.931 |
| | TestIACC | 0.955 | 0.958 | 0.953 | 0.914 | 0.965 | 0.886 | 0.881 | 0.911 | 0.897 | 0.950 |
| | TestIIACC | 0.963 | 0.928 | 0.965 | 0.851 | 0.947 | 0.811 | 0.816 | 0.901 | 0.835 | 0.883 |

*Note. The abbreviation of preTrainACC and preValACC represents the pre-training ACC of the training and validation sets; TrainACC, ValACC, TestIACC, and TestIIACC stand for the ACC predicted by the models on the training, validation, test I and test II sets.

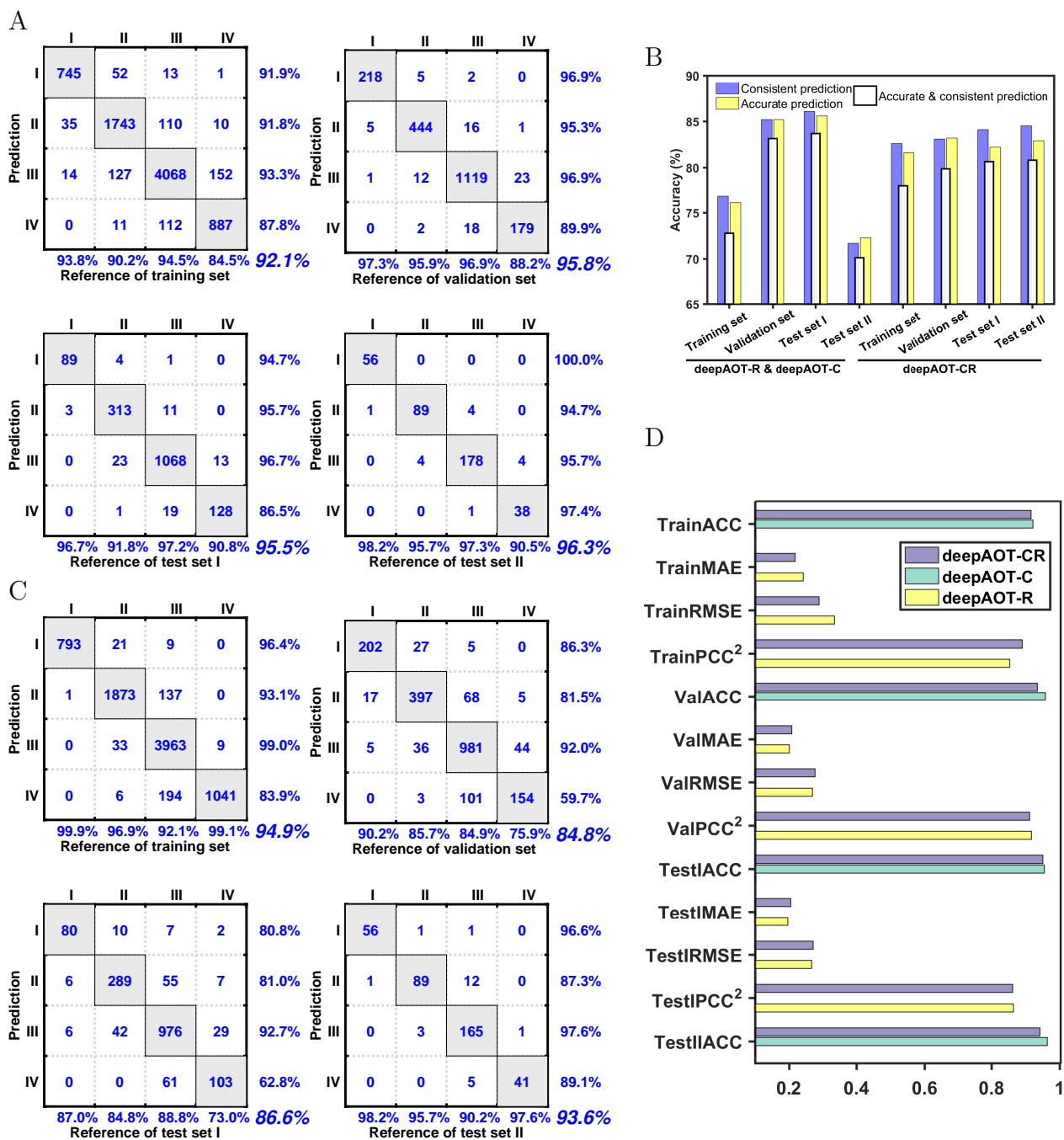

Figure 3: (A) The confusion matrix of deepAOT-C. (B) Consistency comparison between deepAOT-R & deepAOT-C and deepAOT-CR. (C) The confusion matrix for SVM_CM1, which is a SVM model with deep fingerprints from CM1. (D) Performance comparison of deepAOT-CR, deepAOT-R and deepAOT-C.TrainACC, ValACC, TestIACC and Test‖ACC mean the ACC index of the training, validation, test I and test II set, respectively. Different suffix represents different indicators.

## Performance Evaluation of MCMs

The MCM, as a semi-quantitative description for AOT, is more intuitive in toxicity estimation than the more simplistic numbers predicted by RMs, which creates difficulty in understanding chemical toxicity.

In order to develop high-level MCMs, the 500 random sets of hyper-parameters were set in the MGE-CNN, as were the 500 models with different topological networks that were pre-trained with 1000 iterations. After pre-training, the top 10 models were selected with the highest ACC of the validation set (Table 2). Of these, different sets of hyper-parameters resulted in large differences on ACC of the validation set (83.9-94.2%). After the next 1000 iterations were finished, all of the 10 sets of well-trained weights were selected and stored for external predictions. The satisfactory results are displayed in the rows of "TrainACC", "ValACC", "TestIACC" and "TestIIACC" of Table 2. Of these values, ACC in the validation set was between 86.6-96.3%, while the ACC range for the two test sets were from 81.1-96.5%. The CM1 (deepAOT-C) had the best external prediction ability (with fewer feature dimension) for test set I (ACC of 95.5%) and test set II (ACC of 96.3%) among the 10 models. The confusion matrix of deepAOT-C is portrayed in Figure 3A. The high Sens and PPV index for each class and the high ACC demonstrated that deepAOT-C performed better than the previously reported MCM of Li et al.[15] for the validation set (ACC of 83.2%) and the two external test sets (ACC of 83.0%, ACC 0f 89.9%, respectively). These data indicate that deepAOT-C has an excellent generalization ability. In addition, it is suggestive that the MGE-CNN architecture could be successfully extended to multi-classification problems.

## Performance Evaluation of Multi-task Models

The multi-task deepAOT-CR model was constructed with the hyperparameters of deepAOT-C. The modified cost function is as follows.

$$J(\theta) = J_C(\theta) + \beta J_R(\theta) + \alpha \|\theta\|_2$$

Here, $J_C(\theta)$, $J_R(\theta)$ is the loss of classification task and regression task, respectively. $\beta \in (0, 1]$ is a weight parameter to be trained with a smaller learning rate. The comparable performance of deepAOT-CR with that of deepAOT-C and deepAOT-R is shown in Figure 3D and Figure S5. Although it is slightly lower than the single-task deepAOT-C and deepAOT-R, deepAOT-CR was demonstrated to outperform each of all the single models (shown in Table S2) for regression task. More importantly, it could be used for simultaneous predictions of the classification and regression tasks, which suggested that it was appropriate for the MGE-CNN architecture to achieve multi-task problems.

## Consistency Analysis of RMs and MCMs

In order to examine the consistency between RMs and MCMs, the deepAOT-R and deepAOT-C were analyzed together. The outcomes of deepAOT-R were assigned to the category space. The consistent prediction outcomes of both models was counted for each data set (Figure 3B). For the consistent prediction, the percentages on the four data sets were 76.8%, 85.2%, 86.1%, 71.7%, respectively. The accurate classification prediction of deepAOT-R was 76.1%, 85.2%, 85.6% and 72.3%, respectively. Meanwhile, the consistent and accurate predictions respectively occupied 72.8%, 83.2%, 83.7%, 70.1% for each data set. Such comparisons suggested that most of the consistent predictions were corresponded to correct labels, which meant there was a high consistency between the deepAOT-R and deepAOT-C. For the deepAOT-CR, the consistent outcomes of regression and classification were 82.6%, 83.1%, 84.1%, 84.5%, respectively, which improve the overall consistency for the four data sets. The consistent and accurate predictions respectively occupied 77.9%, 79.9%, 80.6%, 80.8% for each data set, shown in Figure 3B. From the view of Figure 2C (green) and Figure S6, deepAOT-CR could significantly (p-value of paired t-test < 0.001) improve the distinguishing capability for category IV.

## Forward Exploration of Fingerprints

The forward exploration evaluated the extent by which the fingerprints from the MGE-CNN-based models favored of shallow decision making systems, such as MLR and SVM. For this purpose, fingerprints were extracted from the "Fingerprint" layer in the well-trained deep models, then the whole data set was transferred into a matrix of $N$ (number of compounds) × FPL, which was a featurization and vectorization process for compounds. This operation was executed for both RMs and MCMs. For the RM4, the matrix for the training set, 8080 (compounds) × 48 (features), was regarded as an input for MLR, fitting the target values of $\log(LD_{50})$ by minimizing the sum of the squares of the vertical deviations from each data point to the best-fitting line. The best-fitting line for the training set was calculated, and was used to predict the validation and test I sets (total of 3718 compounds). Performance of the MLR models with deep fingerprints are summarized in Table 3. In which, the MAE, RMSE and $PCC^2$ were calculated for the training set and the validation and test I sets. The MAE and RMSE range for the validation and test I sets was from 0.378-0.427 and from 0.499-0.561, respectively, while the $PCC^2$ was in the range of 0.554-0.650. The consensus model also demonstrated significant improvement for the training and external test sets, and the performance of MAE, RMSE and $PCC^2$ was 0.348, 0.465 and 0.696, respectively. These prediction levels are completely acceptable for a MLR method. When the LLR (which was an improved MLR method) reported by Lu *et al.*[16] was challenged by "Set_3874", the $PCC^2$ and MAE of the consensus model (with different molecular fingerprints: ECFP4, FCFP4,[24] MACCS, and physicochemical descriptors from commercial software[53,54]) were 0.608 and 0.420, respectively (Figure S7). A pure MLR method based on deep fingerprints was used to ensure that $PCC^2$ and MAE would stay in a range of 0.554-0.650 and 0.378-0.427, respectively. Comparing the two, whether for a single model or the consensus model, the MLR models outperformed LLR models at a similar level test set size, which revealed that deep fingerprints were more useful than application-specific molecule descriptors or fingerprints for AOT prediction without an idea of "Clustering first, and then modelling".[55]

Table 3: Performance of MLR models with deep fingerprints from MGE-CNN architecture on the training, validation, and test I sets.

| Model [*] | Evaluation index [*] | | | | | |
|---|---|---|---|---|---|---|
| | TrainMAE | TrainRMSE | TrainPCC$^2$ | Val&TestIMAE | Val&TestIRMSE | Val&TestIPCC$^2$ |
| MLR_RM1 | 0.428 | 0.558 | 0.580 | 0.404 | 0.542 | 0.584 |
| MLR_RM2 | 0.425 | 0.556 | 0.583 | 0.402 | 0.538 | 0.592 |
| MLR_RM3 | 0.418 | 0.544 | 0.600 | 0.403 | 0.528 | 0.606 |
| MLR_RM4 | 0.432 | 0.563 | 0.572 | 0.405 | 0.535 | 0.596 |
| MLR_RM5 | 0.410 | 0.538 | 0.610 | 0.397 | 0.524 | 0.614 |
| MLR_RM6 | 0.442 | 0.578 | 0.549 | 0.427 | 0.561 | 0.554 |
| MLR_RM7 | 0.437 | 0.569 | 0.563 | 0.415 | 0.553 | 0.566 |
| MLR_RM8 | 0.402 | 0.523 | 0.631 | 0.378 | 0.499 | 0.650 |
| MLR_RM9 | 0.422 | 0.548 | 0.595 | 0.398 | 0.521 | 0.614 |
| MLR_RM10 | 0.432 | 0.561 | 0.575 | 0.414 | 0.542 | 0.583 |
| Consensus | 0.379 | 0.497 | 0.679 | 0.348 | 0.465 | 0.696 |

[*]Note. MLR_RM$i$ means the MLR model constructed by deep fingerprints from RM$i$, $i \in \{\{1, 2, ..., 10\}\}$. "Consensus" means the average outcomes of the above 10 models. Val&TestIMAE, Val&TestIRMSE and Val&TestIPCC$^2$ are MAE, RMSE and PCC$^2$ of the merged validation and Test I set.

For the MCMs, fingerprints were also extracted, and the training part was used to construct multi-class SVM$_{OAO}$) models with the "scikitlearn" package[56] in Python 2.7. The Gaussian radial basis function kernel was used and the parameters $C$ and $\gamma$ were tuned with the validation set. The performance of SVM$_{OAO}$ models with deep fingerprints was assessed with ACC index (Table 4). The range for the training set was from 84.2-96.5% and the validation range was between 78.7-84.8%. For the two external sets, an acceptable ACC range is from 77.9-94.9%. Among the SVM models, SVM_CM1 had the best ACC of 94.9% for the training, 84.8% for the validation set, 86.6% for test set I and 93.6% for test set II. Meanwhile, the confusion matrix for SVM_CM1 indicated that the three indexes of SVM_CM1 were better than those of SVM models developed by Li *et al.*,[15] shown in Figure 3C and S8. Therefore, deep fingerprints from MGE-CNN-based RMs and MCMs were better than standard fingerprints, which further demonstrated that the MGE-CNN implemented better MRs for AOT prediction with automatic feature extraction through supervised learning. With analysis of tanimoto distance, Table 5 suggested that deep fingerprints had a high correlation to molecular topological structure-based ECFP4, FCFP4 and MACCS fin-gerprints and were different from randomly generated fingerprints. To a certain extent, it revealed the interpretability and rationality of these deep fingerprints.

Table 4: Performance of SVM$_{\text{OAO}}$ models with deep fingerprints from MGE-CNN architecture.

| Model[*] | Evaluation index | | | |
|---|---|---|---|---|
| | TrainACC | ValACC | TestIACC | TestIIACC |
| SVM_CM1 | 0.949 | 0.848 | 0.866 | 0.936 |
| SVM_CM2 | 0.934 | 0.816 | 0.799 | 0.880 |
| SVM_CM3 | 0.950 | 0.840 | 0.849 | 0.912 |
| SVM_CM4 | 0.959 | 0.800 | 0.793 | 0.885 |
| SVM_CM5 | 0.958 | 0.839 | 0.830 | 0.928 |
| SVM_CM6 | 0.942 | 0.812 | 0.806 | 0.891 |
| SVM_CM7 | 0.857 | 0.787 | 0.779 | 0.840 |
| SVM_CM8 | 0.988 | 0.809 | 0.805 | 0.949 |
| SVM_CM9 | 0.847 | 0.801 | 0.780 | 0.837 |
| SVM_CM10 | 0.965 | 0.791 | 0.780 | 0.909 |

[*] Note. SVM_CM$i$ means the SVM$_{\text{OAO}}$ model based on deep fingerprints from CM$i$, $i \in \{\{1, 2, ..., 10\}\}$.

Table 5: Correlation analysis of tanimoto distance between different fingerprints.

| Fingerprint | Random | ECFP4 | FCFP4 | MACCS | DeepAOT | PCC |
|---|---|---|---|---|---|---|
| Correlation of Tanimoto distance | + | + | | | | 0.193±0.014 |
| | + | | + | | | 0.264±0.030 |
| | + | | | + | | 0.237±0.021 |
| | + | | | | + | **0.212±0.030** |
| | | + | + | | | 0.984 |
| | | + | | + | | 0.952 |
| | | + | | | + | **0.845** |
| | | | + | + | | 0.946 |
| | | | + | | + | **0.834** |
| | | | | + | + | **0.867** |

## Backward Exploration of Fingerprints

The backward exploration of the "Fingerprint" layer was expected to provide an understanding of fingerprint activation.

Herein, only the above RM4 and CM1 was further examined. After linear regression, the most negative correlation feature of the fingerprints was calculated, which represented the most toxic feature. Comparing activation values of this feature, nine values were determined to contribute most to feature activation. The nine values could be mapped into different substructures, thereby suggesting that these substructures were the most correlative to the explored toxicity feature (Figure 4A and 4B). There were mainly two classes of highlighted fragments, $\alpha,\beta$-Unsaturated nitriles (TA626) and alyl (thio)phosphates (TA776) for RM4, while TA776 and thicarbonyl (TA374) for CM1. The three fragments have been reported to be toxicity structural alerts.[57–59]

Further analysis of RM4 and CM1 explored the highlighted fragments for each compound in category I, followed by the approach demonstrated in Figure S9. Table 6 and Table S3 describe the highlighted fragments which RM4 automatically generate. Moreover, we found most of the highlighted fragments could correspond to the reported TAs. For example, some of the corresponding reported TAs were TA392, TA1285, TA777, TA2958, TA890, TA879, TA462, TA583, TA2795, TA1792, TA623, TA1801, TA279, TA260, TA623, TA751, TA584, TA312, TA1938, TA374, TA626, TA362. Only a few of highlighted fragments did not correspond to the reasonable TAs, such as TA660, TA580, TA249. Due to the high consistency with the reported TAs, this approach had potential for inferring TAs for unknown compounds. For CM1, the highlighted fragments of each compound was almost similar to that from RM4, part of which shown in Figure 4C (in which some inconsistent highlighted fragments are also presented). Therefore, besides of AOT prediction, MGE-CNN-based models was also able to infer TAs with analysis of internal activations.

## Conclusion

In this study, RMs and MCMs constructed by the MGE-CNN were used to estimate AOT in rat for chemical safety assessment. The consensus deepAOT-R model had an outstanding performance with higher $PCC^2$ (0.864), lower RMSE (0.268) and lower MAE (0.195) than the previous best models. When using the deepAOT-R to predict toxicity category, the per-formance was within an acceptable level and the recommended range (within 1-fold RMSE) may be important for prediction outcomes. In addition, the deepAOT-C also demonstrated an excellent performance (ACC of validation set: 92.1%; ACC of test set I: 95.8%; ACC of test set II: 96.3%) when compared to the best reported MCMs . In addition, the multi-task deepAOT-CR also presented a high predictive power for simultaneous assessment of regression and classification problems.

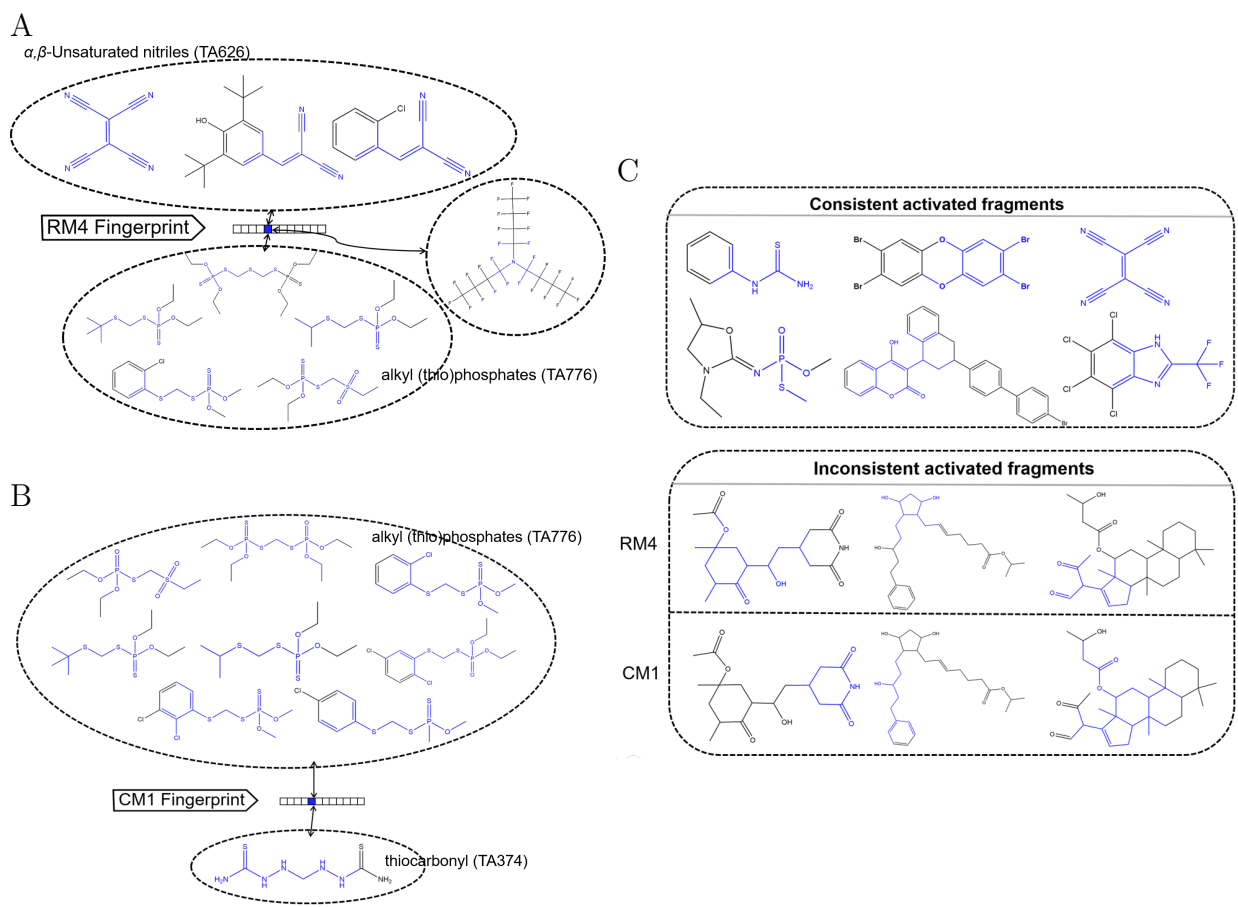

Figure 4: Overview of highlighted fragments. (A) and (B) are Corresponding highlighted fragments that match the most toxic features (blue) of the RM4 and CM1 fingerprint. TA626, TA776, TA374 are the registered numbers from the Online Chemical Database. (C) Consistency comparison of part of the highlighted (blue) fragments for RM4 and CM1.

Table 6: Comparison of TAs and activity fragments inferred by RM4.

| No. | Activation Fragment | Structural Alert | | Alert ID | Reference |
|-----|---------------------|------------------|--|----------|-----------|
| 1 | 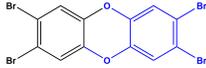 | 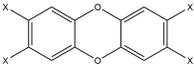 | X = Br, Cl, F, I;<br>Endpoint:<br>Non-genotoxic carcinogenicity | TA392 | ref.[59] |
| 2 | 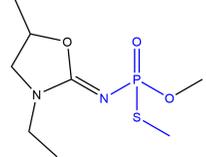 | 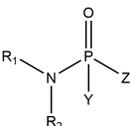 | R1, R2 = H, alkyl, aryl;<br>Y, Z = any O, N, Hal residue;<br>Endpoint:<br>Extended Functional Groups | TA1285 | Checkmol |
| 3 | 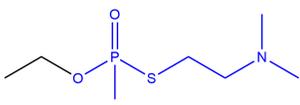 | 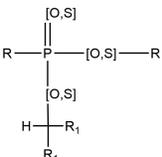 | R = any carbon atom;<br>R1 = H or any carbon atom;<br>Endpoint:<br>Potential electrophilic agents | TA777 | ref.[58] |
| 4 | 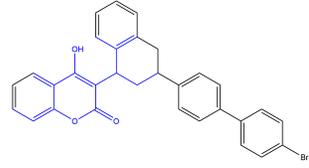 | 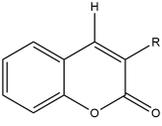 | R = -C=O,-C=S,-S=O,<br>-CN,Hal,-C(Hal)(Hal)(Hal);<br>R ≠ -COOH;<br>Endpoint:<br>Reactive, unstable, toxic | TA890 | ref.[60] |
| 5 | 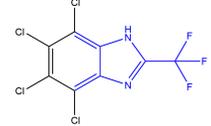 | 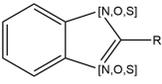 | R = C, N, O, S , Ar;<br>Ar = any aromatic atom;<br>Endpoint:<br>Skin sensitization | TA583 | ref.[61] |
| 6 | 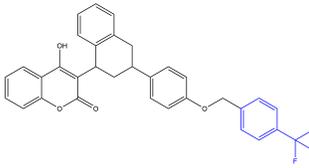 | 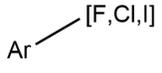 | Ar = any aromatic atom;<br>Endpoint:<br>Nonbiodegradable compounds | TA2795 | ref.[62] |
| 7 | 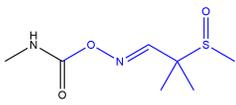 | 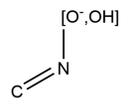 | Linkage = 'HIGH'<br>(PRI = 1.46, PSI = 4.38);<br>Endpoint:<br>Promiscuity | TA1792 | ref.[63] |
| | | 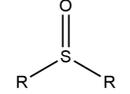 | R = any atom;<br>Endpoint:<br>Acute Aquatic Toxicity | TA623 | ref.[57] |
| 8 | 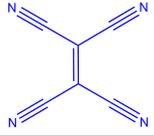 | 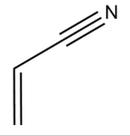 | Endpoint:<br>Acute Aquatic Toxicity | TA626 | ref.[57] |

In view of high-level prediction models, more attention was focused on exploring and interpreting our models. In fact, deep fingerprints extracted from the RMs and the MCMs were able to better support the shallow decision making systems than application-specific molecular descriptors or fingerprints. The consensus MLR model based on these deep fingerprints had a high $PCC^2$ (0.696) and a low MAE (0.348) for the large external set (3718 compounds). Meanwhile, the best SVM model with deep fingerprints also performed very well, with ValACC of 84.8%, TestIACC of 86.6% and TestIIACC of 93.6%. With correlation analysis of tanimoto distance, we recognized that these deep fingerprints were highly correlated with topological structure-based fingerprints. The successes of deep fingerprints could potentially be applied to other tasks related to AOT. One toxicity-related feature of these fingerprints was tracked back to the atomic level and the highlighted toxicity fragments inferred by RM4 and CM1 were compared with the reported TAs. This surprising consistency suggests that the well-trained deep models are no longer "black" models and that these deep models advanced in AOT-related knowledge such that they can be used to infer TAs. Without prior knowledge about fragments, only the information of atoms and bonds can be used to form the knowledge of fragments, all of which are due to the ability of automatic feature learning from deep learning.

The MGE-CNN is not limited to AOT and it could be applied for studying other endpoints induced by compounds in complex systems. Without understanding any mechanism, end-to-end (SMILES-to-endpoint) learning based on a known large data set with high quality can be useful in predicting this endpoint, extracting the endpoint-related fingerprints and inferring the endpoint-related fragments. This methodology is a promising tool in developing and better understanding chemical information of compounds.

# ASSOCIATED CONTENT

Supporting Information Available is available free of charge at http://pubs.acs.org.

# AUTHOR INFORMATION

**Corresponding Authors**

\* To whom correspondence should be addressed.

Fax: (+86)10-62759595, E-mail: jfpei@pku.edu.cn;

Fax: (+86)10-62751725, E-mail: lhlai@pku.edu.cn.

**Notes**

The authors declare no competing financial interest.

# Acknowledgement

The authors thank Prof. Yun Tang, from School of Pharmacy, East China University of Science and Technology for valuable dataset of rat oral $LD_{50}$. The authors also thank David Kristjanson Duvenaud for providing effective parameter ranges. The work was partially carried out at Peking University High Performance Computing Platform, and the calculations were performed on CLS-HPC. This research was supported, in part, by the Ministry of Science and Technology of China (grant numbers: 2016YFA0502303, 2015CB910302) and the National Natural Science Foundation of China (grant numbers: 21673010, 21633001).

## Supporting Information Available

The following files are available free of charge.

Table S1: The set range of some important hyper-parameters.

Table S2: Hyper-parameters and performance of the top 10 RMs with the lowest RMSE of validation set.

Table S3: Additional comparison of TAs and activity fragments inferred by RM4.

Figure S1: Workflow of chemical data curation. RVA is the explicit valence analysis based on the RDKit package; TEV (FEV) set: true (false) explicit valence set; "[N] → [N+]" in the SMILES string indicates that the nitrogen with brackets should be charged; "Structure Checker" module is used to correct the molecular structure; "Standardizer" module was utilized to transform, standardize and unify the canonical SMILES strings.

Figure S2: Plot of the RMSE (left) and PCC (right) from 500 models based on the training (blue) and validation sets (red).

Figure S3: The distribution of prediction errors from the training, validation and test I sets based on deepAOT-R.

Figure S4: The confusion matrix of all the four data sets predicted by deepAOT-R (supplemented with wiggle room of 1-fold RMSE).

Figure S5: The confusion matrix of all the four data sets predicted by classification task of deepAOT-CR.

Figure S6: The confusion matrix of all the four data sets predicted by regreesion task of deepAOT-CR.

Figure S7: Comparison between MLR models with deep fingerprints and LLR models with different standard features (ECFP4, FCFP4, MACCS, etc).

Figure S8: Comparison of ACC for the MGE-CNN-based model, the $SVM_{OAO}$ model with deep fingerprints and the $SVM_{OAO}$ model with MACCS fingerprints.

Figure S9: Schematic diagram for exploring toxicity fragments of flocoumafen. The blue arrows represent well-trained weights. The toxicity (in blue) feature from fingerprints of flocoumafen was traced back into different fragments in different layers, displayed in pink arrows. Comparing all the activation of central atoms for these fragments, the maximum activation fragment is represented in blue (left). This fragment is referred as a toxicity fragment inferred by RM4.